\documentclass[lettersize,journal]{IEEEtran}
\pdfoutput = 1

\usepackage{hyperref}
\usepackage{multirow}
\usepackage{amsmath,amsfonts}
\usepackage{booktabs}
\usepackage{algorithmic}
\usepackage{algorithm}
\usepackage{array}
\usepackage[caption=false,font=normalsize,labelfont=sf,textfont=sf]{subfig}
\usepackage{textcomp}
\usepackage{stfloats}
\usepackage{url}
\usepackage{verbatim}
\usepackage{cite}
\usepackage{color}
\usepackage{balance}
\usepackage{threeparttable}
\usepackage{hyperref}

\usepackage{pdfpages}
\usepackage[dvipsnames]{xcolor}

\usepackage{graphicx}
\usepackage{float}

\begin{document}

\title{Adjacent-level Feature Cross-Fusion with 3-D CNN for Remote Sensing Image Change Detection}

\author{Yuanxin Ye, \textit{Member, IEEE}, Mengmeng Wang, Liang Zhou, Guangyang Lei,
	Jianwei Fan, \textit{and} Yao Qin
\thanks{This paper is supported by the National Natural Science Foundation of China
	(No.42271446 and No.41971281), and in part by the Natural Science
	Foundation of Sichuan Province under Grant 2022NSFSC0537.
	(\textit{Corresponding author: Mengmeng Wang})}
\thanks{Y. Ye, M. Wang, L. Zhou, and G. Lei are with the Faculty of Geosciences
	and Environmental Engineering, Southwest Jiaotong University, Chengdu
	611756, China (e-mail: yeyuanxin@home.swjtu.edu.cn; wm$\_$gmail@163.com; zlup@my.swjtu.edu.cn;
	LGY0520@my.swjtu.edu.cn)}
\thanks{J. Fan is with the School of Computer and Information Technology, Xinyang Normal University, Xinyang 464000, China (e-mail: fanjw@xynu.edu.cn)}
\thanks{Y. Qin is with the Northwesten Institute of Nuclear Technology, Xi’an 710025, China (e-mail:tsintuan@163.com)}
}



\maketitle

\begin{abstract}

Deep learning-based change detection (CD) using remote sensing images has received increasing attention in recent years. However, how to effectively extract and fuse the deep features of bi-temporal images for improving the accuracy of CD is still a challenge. To address that, a novel adjacent-level feature fusion network with 3D convolution (named AFCF3D-Net) is proposed in this article. First, through the inner fusion property of 3D convolution, we design a new feature fusion way that can simultaneously extract and fuse the feature information from bi-temporal images. Then, to alleviate the semantic gap between low-level features and high-level features, we propose an adjacent-level feature cross-fusion (AFCF) module to aggregate complementary feature information between the adjacent levels. Furthermore, the full-scale skip connection strategy is introduced to improve the capability of pixel-wise prediction and the compactness of changed objects in the results. Finally, the proposed AFCF3D-Net has been validated on the three challenging remote sensing CD datasets: the Wuhan building dataset (WHU-CD), the LEVIR building dataset (LEVIR-CD), and the Sun Yat-Sen University dataset (SYSU-CD). The results of quantitative analysis and qualitative comparison demonstrate that the proposed AFCF3D-Net achieves better performance compared to other state-of-the-art methods. The code for this work is available at \href{https://github.com/wm-Githuber/AFCF3D-Net}{https://github.com/wm-Githuber/AFCF3D-Net}
\end{abstract}


\section{Introduction}
\IEEEPARstart{C}{hange} detection (CD) is the process of identifying and extracting change information by comparing bi-temporal images taken in the same geographical area but at different time periods \cite{ref1}.
In other words, two different time images after precise registration \cite{Ye_Registration, Zhu_registration} are used to identify changes on the surface.
CD is a powerful tool that can depict the change of the Earth's surface by generating a binary map \cite{ref2}, and it has been widely utilized in a wide range of remote sensing (RS) applications, including the assessment of natural disasters \cite{ref3}, the management of land \cite{ref4}, and urban expansion \cite{ref5}. Therefore, CD is one of the central issues in RS.

Over the last few decades, a larger number of CD methods have been proposed. Depending on whether they require extracting features manually, the current CD methods can be divided into traditional CD methods and deep learning-based (DL-based) methods \cite{ref6}. In addition, the traditional CD methods can be subdivided into pixel-based CD (PBCD) and object-based CD (OBCD) according to the basic processing unit used \cite{ref2}. The PBCD obtains the change results by directly comparing the individual pixels. The PBCD methods include the algebra-based algorithm (i.e., image difference \cite{ref7} and image rationing \cite{ref8}), the transformation-based algorithm (i.e., change vector analysis \cite{ref9}), and the classification-based algorithm (i.e., post-classification \cite{ref10}).
Despite the fact that PBCD methods are simple and have been widely used, the results are frequently compromised by image noise since the methods only focus on individual pixels and do not take into consideration the spatial context information that exists between neighboring pixels \cite{ref11}.
Researchers introduce the techniques of object-based analysis to CD fields to address the above problems \cite{ref12}. The OBCD methods make use of a set of adjacent local pixels, which are produced through image segmentation utilizing rich features (such as texture, shape, and the spatial relationship) \cite{ref13, ref14, ref15}. In this case, Li \textit{et al.} \cite{ref16} proposed an object-based change vector analysis (OCVA) algorithm for CD. Liu \textit{et al.} \cite{ref17} proposed a CD algorithm by using spatial and shape features.
However, the misclassification error that results from the indeterminacy of the objects in the segmentation results limits the accuracy of the results \cite{ref18}.
Additionally, both the OBCD and PBCD require a significant amount of artificial interference and are highly susceptible to changes in lighting conditions \cite{ref19}.

\begin{figure}[ht]		
	\centering
	\includegraphics[scale=0.85]{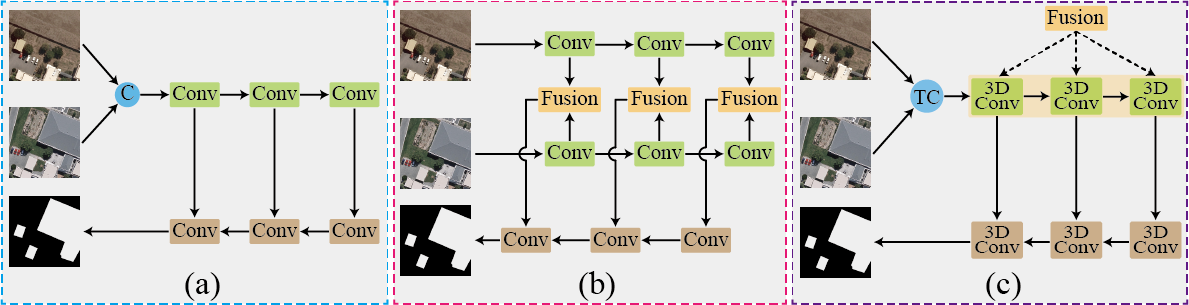}
	\caption{Structure of different fusion strategy. (a) Early fusion structure. (b) Late fusion structure. (c) Proposed fusion structure.}
	\label{figure1}
\end{figure}

With the fast advancement of deep learning (DL) technology \cite{Ye_matching}, especially deep convolutional neural networks (CNNs), which are introduced into CD fields because of their excellent ability for strong multilevel feature extraction and superior performance over traditional methods \cite{ref20}. Compared to the traditional PBCD and OBCD methods,
methods based on CNNs not only decrease the articial work, but they also avoid errors in pre-processing \cite{ref11}. Due to the end-to-end advantages of fully convolutional networks (FCN) \cite{Ye_matching}, many FCN-based CD networks have been proposed \cite{ref21, ref22, ref23}. 
According to the fusion strategy, FCN-based CD networks can be roughly split into early fusion (i.e., single-stream) networks and late fusion (i.e., two-stream) networks. In the early fusion methods,
bi-temporal images are concatenated and then inputted into the network to generate the result,
as shown in Fig. \ref{figure1}(a). For instance, Daudt \textit{et al.} \cite{ref22} first concatenated bi-temporal images at the channel dimension. After that, the newly concatenated data is fed into a U-Net network to detect image changes. Similarity, bi-temporal images are merged into one data by Peng \textit{et al.} \cite{ref21} which is then transmitted into the UNet++ network to produce a change map. Peng \textit{et al.} \cite{ref11} designed a difference-enhancement network that extracts accurate change information by concatenating bi-temporal images. Although early fusion networks have been significantly advanced by deep learning techniques, these methods still face a few problems.
For example, early fusion methods could result in the semantic information of bi-temporal features becoming confusing.
It is challenging to extract information about a single feature from the fused image pairs \cite{ref18}. Different from early fusion methods, late fusion methods first employ a Siamese network to extract dual-branch features from two independent branches \cite{ref24}. After that, the extracted bi-temporal features are fused at each level via some fusion modules. As shown in Fig. \ref{figure1}(b). Daudt \textit{et al.} \cite{ref22} applied the Siamese network to fuse bi-temporal features by concatenation or difference for CD tasks. Since then, the idea has been widely adopted and has become the baseline network for CD \cite{ref25, ref26, LuoROC}. Based on the above-mentioned analysis, DL-based methods have achieved promising results, but the current DL-based methods still have much room for improvement.

Motivated by the superior ability of 3D CNNs to simultaneously extract and fuse spatial-temporal features within video frames \cite{ref27, ref28}, this paper proposes a new feature fusion strategy [Fig. \ref{figure1}(c)] and a novel CD network based on 3D CNNs.
We reconsider the bi-temporal images as the before and after frames of a video, in a manner analogous to the way that 3D convolution is used to extract spatial-temporal features of video frames. Specifically, we introduce 3D convolution to extract the bi-temporal features, where the extracted bi-temporal features have been fused based on the inner fusion property of 3D convolution. In addition, to alleviate the gap between high-level semantic features and low-level spatial features, we propose an adjacent-level feature cross-fusion (AFCF) module to explore the complementary information contained in adjacent-level features. Finally, after the adjacent-level features are merged by the AFCF, the decoding part adopts the full-scale skip connected strategy to produce change results. To sum up, the contributions of this paper are summarized as follows:
\begin{enumerate}
	\item {A novel end-to-end CD method based on 3D convolution is proposed for RS images. Different from the existing feature fusion strategy, the proposed method is an innovative attempt in the CD task to use the inner fusion property of 3D convolution.}
	\item {An adjacent-level feature cross-fusion (AFCF) module is designed. AFCF can effectively alleviate the gap between high-level semantic features and low-level spatial features to assist the network in obtaining more accurate change results.}
	\item {To demonstrate that the proposed method is effective, extensive experiments have been conducted on three challenging datasets. The proposed network reaches an F1-score of 93.58\%, 90.76\%, and 83.11\% on the WHU-CD, LEVIR-CD, and SYSU-CD datasets, respectively. The quantitative results show that the proposed network significantly outperforms several state-of-the-art (SOTA) DL-based methods.}
\end{enumerate}

The rest of this paper is organized as follows: In Section II,
we will introduce some related works.
The specifics of the proposed method will be described in Section III. In Section IV, several experiments on different datasets will be conducted to validate the proposed model. Finally, Section V presents the conclusions.

\section{Related Works}
In this section, we mainly talk about DL-based CD methods. CNN-based methods have proven successful in the image processing field because CNNs can automatically extract hierarchical, non-linear, and complicated features from raw data \cite{ref2}. Accordingly, many studies focus on CNN-based CD methods.

Most of the current CNN-based CD methods use the FCN. This is because the FCN architecture can not only reduce the network parameters but also conduct an end-to-end pixel-wise prediction \cite{ref29}. Daudt \textit{et al.} \cite{ref22} are among the earliest researchers who applied the FCN to CD tasks. They come up with three different UNet-based versions of the FCN: FC-EF, FC-Siam-diff, and FC-Siam-conc. FC-EF fed the network with concatenated pairs of images, while FC-Siam-diff and FC-Siam-conc used the Siamese network to extract the bi-temporal features. Since then, UNet and Siamese structures have emerged as the benchmark models for CD. For instance, Khalid \textit{et al.} \cite{ref30} designed a CD algorithm in which the Siamese structures are modified with the UNet.
A CD network is proposed by Zhang \textit{et al.} \cite{ref24}, which utilizes the Siamese structure to extract representative deep features,
these features are then fed into a difference discrimination network to generate change maps. Similarly, Mesquita \textit{et al.} \cite{ref31} proposed an autoencoder method for CD; where the method reduced the number of labeled samples necessary by utilizing an autoencoder. A Siamese convolutional network with a dual-task constraint is proposed by Liu \textit{et al.} \cite{ref32} as a way to simultaneously complete the tasks of CD and semantic segmentation. Despite the fact that the skip connection based on UNet can improve the performance of CD, the accuracy is still not satisfactory since RS images involve more complicated information and higher resolution \cite{ref33}. Therefore, following the UNet’s skip connection design, some researchers proposed a series of variant extensions of the UNet, such as the UNet++ and the UNet3+. Peng \textit{et al.} \cite{ref21} proposed a network by replacing UNet with UNet++, which achieves better performance for CD. Fang \textit{et al.} \cite{ref34} advocated using densely connected Siamese structures that are based on UNet++ to fuse features that are at different scales.

The attention mechanisms originate from computer vision, and it has been shown that they have a lot of potential to highlight important information that can be used for many image interpretation tasks \cite{ref12}. As a result, researchers have introduced it into CD \cite{ref12, ref35}. For example, Zhang \textit{et al.} \cite{ref24} utilized the channel and spatial attention mechanisms to put more focus on the changed features while simultaneously suppressing the irrelevant ones. In addition, self-attention has the advantage of using the intrinsic information of features as much as possible for attentional interaction when compared to the spatial attention discussed before \cite{ref19}. As a result, researchers have proposed many improved self-attention networks for CD. For example, an adaptive attention module that combines channel and spatial features has been proposed by Wang \textit{et al.} \cite{ref26}. The goal of this mechanism is to generate connections between different scale changes and generate more accurate results. Chen \textit{et al.} \cite{ref36} proposed a network that takes advantage of a dual-attention module in order to detect areas that have changed and obtain more discriminating feature representations. In short, because of its interpretability and plug-and-plug property, the attention mechanism has become a popular research area for CD tasks.

\begin{figure*}[b]		
	\centering
	\includegraphics[scale=0.9]{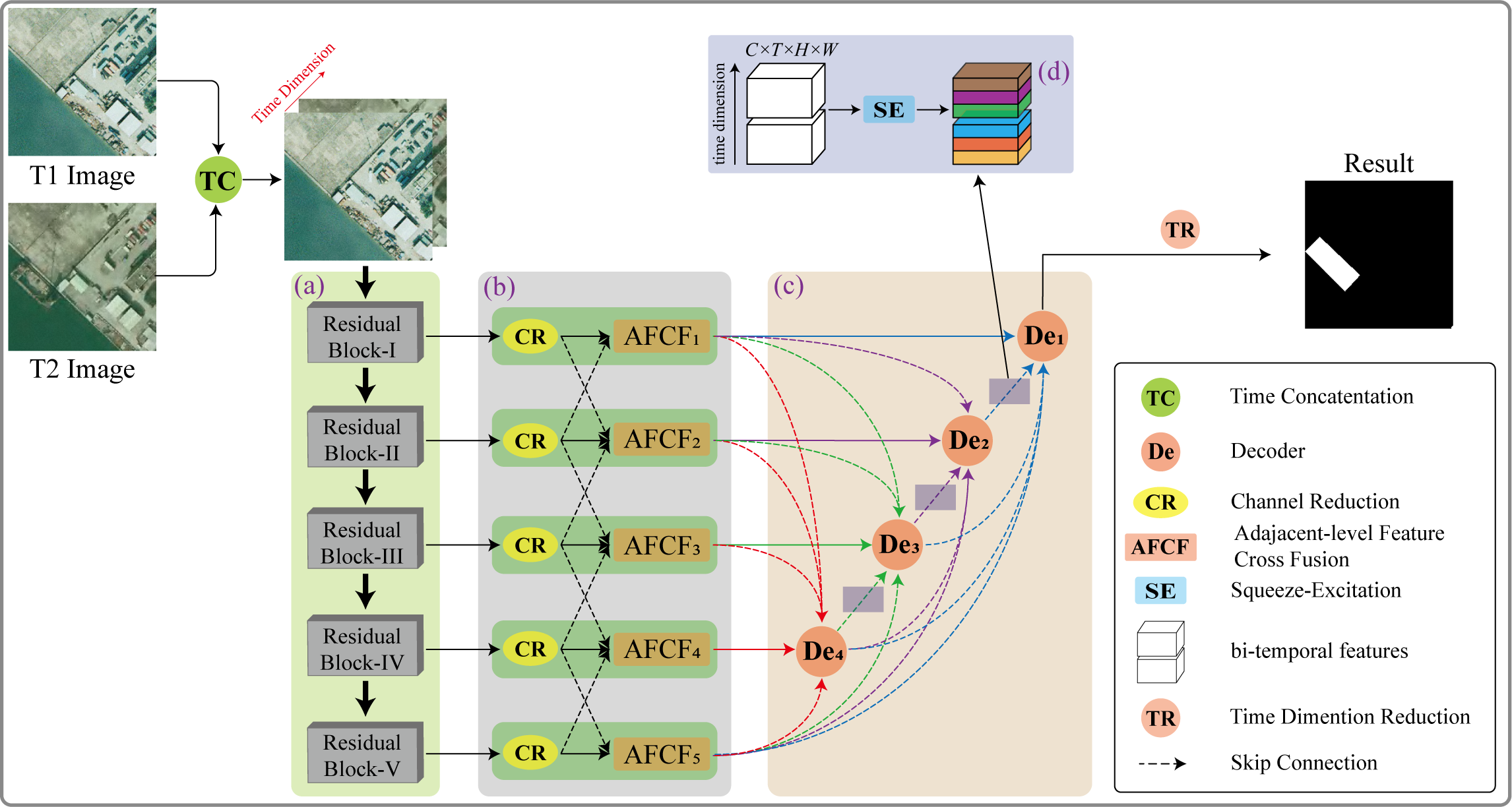}
	\caption{The architecture of the proposed AFCF3D-Net.}
	\label{figure2}
\end{figure*}

Recently, when the Transformer \cite{ref37} is used for the first time in natural language processing (NLP), it showed outstanding results. Researchers gradually applied the Transformer to computer vision tasks like object detection \cite{ref38, object_detection}, image captioning \cite{ref39}, semantic segmentation \cite{ref40, ref41}, and image classification \cite{ref42} because of its strong representation capabilities. Subsequently, the Transformer model has also been applied to CD tasks by researchers. For example, Song \textit{et al.} \cite{ref43} proposed a CD network that makes use of a Transformer to optimize the extracted features and feeds feedback into the original features in the encoder to assist in transforming the pixel space. Mao \textit{et al.} \cite{ref44} proposed a novel CD network through the use of a Transformer. This method makes use of the Transformer to learn the feature intra-relationships at different scales that are extracted by the ResNet, and it then aggregates these features to obtain more accurate semantic and localization representations.

\section{Methodology}
In this section, we first clarify the structure of the AFCF3D-Net in Section III-A. Then, the proposed 3D convolutional block is presented in detail In Section III-B. We give the detailed formulas of our AFCF module in Section III-C. Subsequently, Section III-D provides the decoder part of the AFCF3D-Net. Finally, we introduce the loss function in Section III-E.
\subsection{Overall Structure of the Proposed Network}
Fig. \ref{figure2} shows the architecture of the proposed AFCF3D-Net, which includes three major components:
1) 3D feature encoder: the 3D feature encoder uses a 3D convolution-extended version of the improved ResNet-18 \cite{ref45} as the encoder. In this stage, a group of coarse fusion features can be obtained based on the inner fusion property of 3D convolution. 2) AFCF module: the AFCF module is established between the encoder and decoder and can achieve the cross-fusion of the adjacent-level features from the encoder and deliver valuable change information to the decoder. In addition, we introduce the squeeze-and-excitation module \cite{ref46} and further improve it to fit the 3D convolution. 3) 3D decoder: a variant of the full-scale connection decoder is proposed to achieve a fine fusion of full-scale features. Finally, the fine fusion gradually generates accurate change maps.
\subsection{3D Feature Encoder}
3D CNNs have come to widespread use in a variety of different tasks, such as action recognition \cite{ref47, ref48}, image classification \cite{ref49}, image segmentation \cite{ref50}, and video processing \cite{ref51}. Similar to using 3D CNNs to extract spatial-temporal features of frames of video \cite{ref51}, we reconsider the bi-temporal images as the front and back frames of a video. Inspired by that, this paper introduces 3D convolution to the CD field and proposes the AFCF3D-Net for CD tasks. As mentioned above, the 3D CNNs have been explored many times, they are just like the standard 2D convolution, but with a temporal filter in the time direction. A significant characteristic of 3D CNNs is that they can directly generate hierarchical feature representations of spatial-temporal data. However, a drawback of 3D networks is that they have a significantly larger number of parameters than 2D CNNs do as a result of the additional time dimension. Inspired by Carreira \textit{et al.} \cite{ref52}, the feature encoder is an improved ResNet-18 \cite{ref45} in which we replace all the 2-D convolution kernels of size 3$\times$3 in the conventional ResNet with a 3D convolution kernel of size 3$\times$3$\times$3.

\begin{figure}[!h]		
	\centering
	\includegraphics[scale=0.9]{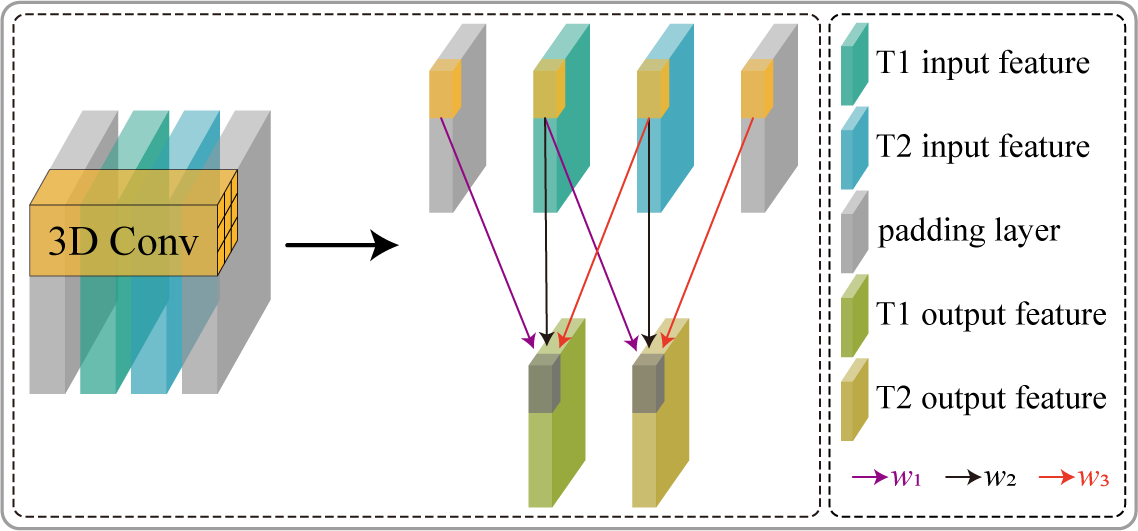}
	\caption{Feature extraction and fusion process of 3D convolution.}
	\label{figure3}
\end{figure}

Given a pair of images $I_1, I_2 \in R^{C\times H\times W}$, where $W$, $H$, and $C$ denote the width, height, and band numbers of the bi-temporal images, respectively. First, the bi-temporal images are jointed together in the time direction to form a 4D tensor, and the 4D tensor can be represented as $C\times T\times H\times W$, where $T$ denotes the time dimension and $T=2$. For convenience, we omit the batch-size. Then, the 4D tensor is fed into the 3D feature encoder to extract multi-level features. Specifically, all the 3D convolution parameters setting for stride, padding, and output dimension in the \textit{T} direction are 1, 1, and 2, respectively. It is essential to point out that the output dimension is the same as the input dimension in the \textit{T} direction. Fig. \ref{figure3} illustrates the 3D convolution process of the 4D tensor in the \textit{T} direction, the features that are output by the 3D encoder can be written as:
\begin{equation}	\label{eq1}
	\begin{cases}
		F_i^1=0 * w_1 + F_{i-1}^1 * w_2 + F_{i-1}^2 * w_3 \\
		F_i^2=F_{i-1}^1 * w_1 + F_{i-1}^2 * w_2 + 0 * w_3
	\end{cases}
\end{equation}
where $*$ denotes the convolution operation, $w_1$, $w_2$, and $w_3$ are the three-layer weight component of the 3D convolution filter, respectively. $F_i^1$ and $F_i^2$, $i\in {0, 1, 2, 3, 4}$ are the output fused features of bi-temporal images, respectively. Specifically, when $i$ is 0, the $F_{i-1}^1$ and $F_{i-1}^2$ are the original images (i.e., $I_1$ and $I_2$). Through the formulation of the 3D convolution above, the bi-temporal features can be effectively fused. In particular, the bi-temporal features have a shared filter weight $w_2$, which can learn the shared representation feature. On the other hand, two individual learnable weights $w_1$ and $w_3$ can provide the difference features to each output features of bi-temporal.

\begin{figure}[ht]		
	\centering
	\includegraphics[scale=0.9]{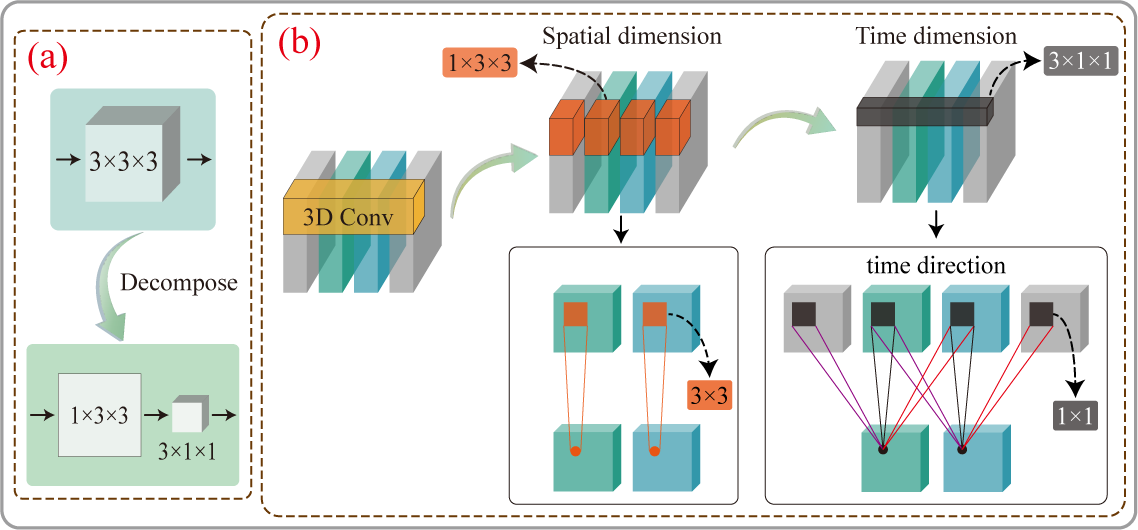}
	\caption{(a) The sketch process of decomposing 3D convolution into a spatial component and a time component. (b) The factorization and encoding process of 3D convolution.}
	\label{figure4}
\end{figure}

To reduce the network’s parameters, we factorize the 3D convolution kernels into separate spatial and time components \cite{ref53}. Specifically, the 3D convolution kernel (3$\times$3$\times$3) can be replaced with a (2+1) D block consisting of a spatial convolution kernel of size 1$\times$3$\times$3 and a time convolution kernel of size 3$\times$1$\times$1, as shown in Fig. \ref{figure4}(a). Since the value of the spatial kernel is 1 in the time dimension, the spatial convolution can be considered a shared filter to extract bi-temporal features. Moreover, it can be noticed that only the time convolution kernel involves the time dimension, so we further factorize the time convolution kernel into three equivalent convolutions with size 1$\times$1. Bi-temporal images share the middle kernel of three convolutions, and the other two boundary kernels belong to one of the bi-temporal images, respectively. After that, the output of the two boundary convolutions is added to the output features on the opposite temporal, and the output features of the bi-temporal are concatenated in the time direction to obtain the fusion features. Feature sizes of five 3D encoding block are 64$\times$2$\times$128$\times$128, 64$\times$2$\times$64$\times$64, 128$\times$2$\times$32$\times$32, 256$\times$2$\times$16$\times$16, and 512$\times$2$\times$8$\times$8, respectively. For convenience, the output features of five encoding block are denoted as $F_o^i, 0\le i\le4$. The above 3D convolution factorization and encoding processes are shown in Fig. \ref{figure4}(b).

\subsection{Adjacent-level Feature Cross-Fusion}
In CNNs, different feature levels of convolutional layers correspond to different feature representations. In particular, the low-level features usually provide spatial details, while the high-level features are rich in semantic information \cite{ref54}.
The multi-level integration can aggregate spatial details and semantic information. However, there is a semantic gap between the low-level and high-level features, and this gap will lead to some inconsistency and confusion for the network when it is constructed by directly concatenating the low- and high-level features \cite{ref55}. Using feature integration between the adjacent-level features is an effective strategy to address the above issues. Accordingly, we have designed the AFCF module that refine both the spatial details and semantic information.

Before the adjacent-level feature cross-fusion, the channel number of each level feature generated from the 3D encoder block is first reduced using a channel reduction operation, which can be described as:
\begin{equation}	\label{eq2}
	\hat{F}_o^i = Conv_{3d}(F_o^i),0\le i \le 4
\end{equation}
where $F_o^i$ is the encoding feature and $\hat{F}_o^i$ is the output feature after channel reduction operation. $Conv_{3d}$ is a 3D convolution operation with a kernel size of $1\times1\times1$, which can reduce the feature channel numbers from $\{64, 64, 128, 256, 512\}$ to $\{32, 32, 32, 32, 32\}$ for each feature level, respectively. By doing this, the channel reduction operation can not only reduce memory usage but also decrease computational load.

\begin{figure}[h]		
	\centering
	\includegraphics[scale=0.9]{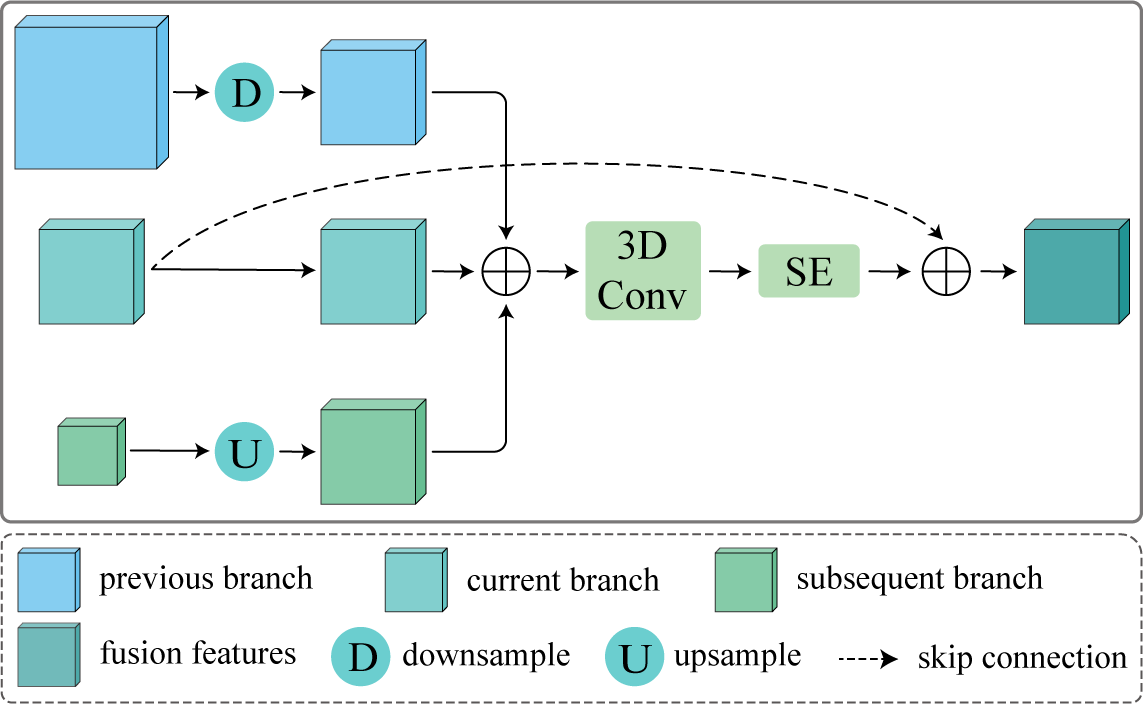}
	\caption{Construction process of AFCF module.}
	\label{figure5}
\end{figure}

Generally, there are three branches (current, previous, and subsequent) in the three middle AFCF module, but there are only two branches (current and one adjacent) in the two boundary AFCF.
Fig. \ref{figure5} shows the structure of the middle-level AFCF. In the first step of cross-fusion, the previous branch (i.e., low-level) and the subsequent branch (i.e., high-level) are transformed to the scale of the current branch by down-sampling and up-sampling, respectively. This procedure can be written as:
\begin{equation}	\label{eq3}
	\begin{cases}
		\bar{F}_o^{i-1}=D(\hat{F}_o^{i-1}),    i=0,1,2,3\\
		\bar{F}_o^{i+1}=U(\hat{F}_o^{i+1}),    i=1,2,3,4\\

	\end{cases}
\end{equation}
where $i$ denotes the $i$-th feature encoding level, $D$ denotes the down-sampling operation with 2 strides, and $U$ is the 2 times up-sampling by bilinear interpolation. $\hat{F}_o^{i-1}$ and $\hat{F}_o^{i+1}$ are the features of the previous branch and the subsequent branch, respectively. $\bar{F}_o^{i-1}$ and $\bar{F}_o^{i+1}$ correspond to the down-sampling and the up-sampling features, respectively. Then, the element-wise addition operation merges both of them into the current branch (i.e., $\hat{F}_o^i$).
Then, the initial cross-fusion features are put into a 3D convolution layer with a kernel of 3$\times$3$\times$3. Meanwhile, the squeeze and excitation (SE) mechanism \cite{ref46} and skip connection strategy \cite{ref45} is introduced to refine the fused features and original current branch features, respectively. The process is formulated as follows:
\begin{equation}	\label{eq4}
	AF^i=
	\begin{cases}
		\hat{F}_o^i+SE(Conv(sum(\hat{F}_o^i,\bar{F}_o^{i+1}))),i=0\\
		\hat{F}_o^i+SE(Conv(sum(\bar{F}_o^{i-1},\hat{F}_o^i,\bar{F}_o^{i+1}))), i=1,2,3\\
		\hat{F}_o^i+SE(Conv(sum(\bar{F}_o^{i-1},\hat{F}_o^i))),i=4\\
	\end{cases}
\end{equation}
where $\bar{F}_o^{i-1}$, $\hat{F}_o^i$, and $\bar{F}_o^{i+1}$ the features of previous branch, current branch, and subsequent branch, respectively. $Conv$ denotes a 3$\times$3$\times$3 convolution. $sum$ is the element-wise summation operation,
$SE$ denotes the SE module introduced below, and $AF^i$ represents the cross-fusion results of adjacent-level features.

\begin{figure}[ht]		
	\centering
	\includegraphics[scale=0.9]{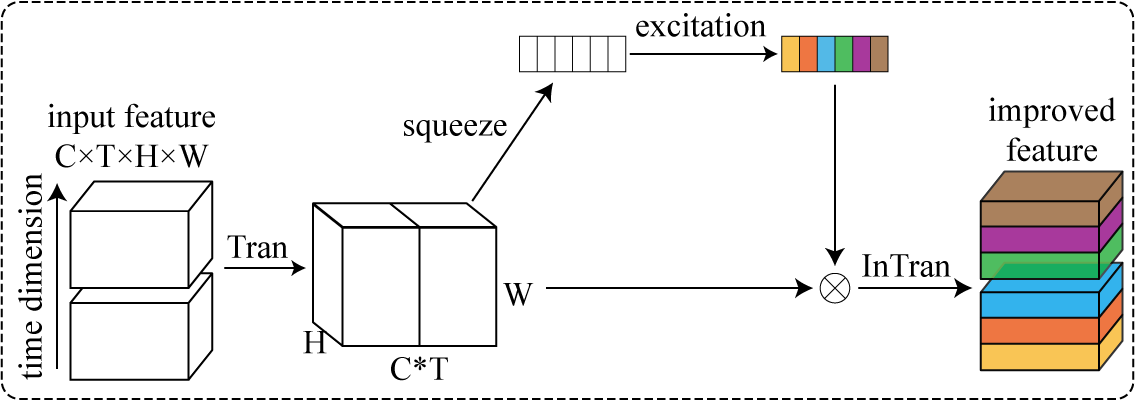}
	\caption{The construction process of the SE module.}
	\label{figure6}
\end{figure}

According to the previous studies \cite{ref56, ref57, ref58}, the SE module has been widely used in CD and achieves better performance. However, the feature size generated from the 3D encoder is a 4D tensor, which contains a time dimension. Accordingly, we improve the SE module to fit the 4D features. The input feature is denoted as $f^{C\times T\times H\times W}$, where $C$, $T$, $H$ and $W$ is the channel numbers, time dimension, height, and weight of the feature, respectively, for convenience, we omit the batch-size. First, the input feature is transformed to the size $(C*T)\times H\times W$ to combine the time dimension information and the channel dimension. Then, the transformed feature is passed through a SE module, which can improve the feature representation. Subsequently, the improved feature is transformed back from $(C*T)\times H\times W$ to $C\times T\times H\times W$. The process is shown in Fig. \ref{figure6} and formulated as follows:
\begin{equation}	\label{eq5}
	F_{se}=Inv(Tr(f)\otimes SE(Tr(f)))
\end{equation}
where $f$ is the feature vector, $Tr$ and $Inv$ denote the forward transform and inverse transform operations, respectively. $\otimes$ represents the element-wise multiplication operation, and $F_{se}$ represents the refined features.
\subsection{Feature Decoder}
The full-scale skip connection has been useful in many fields, such as hyperspectral image classification \cite{ref59}, medical image segmentation \cite{ref60, ref61}, and change detection \cite{ref34}. Compared to the simple skip connection structure of Unet \cite{ref62}, the full-scale skip connection can aggregate the spatial details from low-level features and the abstract semantic from high-level features. Therefore, in order to ensure that the decoder can fully excavate multiscale feature information and thereby improve the accuracy of the results, we make use of the full-scale skip connection strategy, which is shown in Fig. \ref{figure2}(c). On one hand, the semantic gap between the low-level features and the high-level features is alleviated because each level’s features are fused across the AFCF module. On the other hand, different from the previous studies \cite{ref12, ref57}, the bi-temporal features of each level are concatenated in the time dimension, we attempt to fuse the full-scale features in the time direction for CD.

To be more specific, to demonstrate how the feature maps are integrated, we take the 3-th feature decoder as an example. First, the three low-level feature maps (i.e., $AF_o^0$, $AF_o^1$, and $AF_o^2$) with higher resolution are down-sampled using a sequence of down-sampling blocks to the same scale as $AF_o^3$, respectively, while the high-level feature map (i.e., $AF_o^4$) with lower resolution is up-sampled to the same scale as $AF_o^3$. It should be noted that the number of time dimensions for each level feature will remain unchanged. Then, the down-sampled and up-sampled features and $AF_o^3$ are concatenated in the time dimension to generate the concatenated features named $CF^3$. This process can be described as:
\begin{equation}	\label{eq6}
	CF^3=TCat(D(AF_o^0),D(AF_o^1),D(AF_o^2),AF_o^3,U(AF_o^4))
\end{equation}
where $TCat$ denotes the feature concatenation in the time dimension. $U$ and $D$ represent the up-sampling and down-sampling and operation, respectively. $SE$ denotes the SE module which mentioned above.
$CF^3$ is the concatenated features in the time dimension, and the number of time dimensions is 10 (i.e., $T=10$).
Subsequently, the  $CF^3$  is fed into a 3D convolution block to generate the fused features. By performing this operation, we can not only fuse the full-scale features in the time dimension but also reduce the time dimension’s number of fused features to the original to facilitate subsequent processing. The above proves can be described as:
\begin{equation}	\label{eq7}
	F^3=FR(CF^3)
\end{equation}
where $FR$ is a 3D convolution block that contains three consecutive 3D convolutions with 3$\times$3$\times$3 (time stride is 1), $4\times3\times3$ (time stride is 2), and $3\times1\times1$ kernels, respectively. Following a similar procedure, we can obtain the fused features $F^2$, $F^1$, and $F^0$ in sequence. After the $F^0$ is obtained, a filter of $1\times1\times1$ convolution followed by a sigmoid function is applied to produce the change result.
\subsection{Loss Function}
The CD can be viewed as a binary classification task of “unchanged” and “changed”. Therefore, the binary cross-entropy (BCE) loss is used as the loss function and can be written as:
\begin{equation}
		\mathcal{L}_{bce}(t,p)=-\frac{1}{N}\sum_{i=1}^{N}[t_i log(p_i) + (1-t_i) log(1-p_i)
\end{equation}
where $N$ is the total pixel number,
$t$ is the truth map, $t_i$ represents the ground truth of pixel $i$, and $t_i=0$ and $t_i=1$ denote unchanged and changed categories, respectively.
$p$ is the predicted change map, $p_i$ denotes the predicted probabilities of changed pixels, and $1-p_i$ denotes the predicted probabilities of unchanged pixels, $p_i\in[0,1]$.

During the training of the network, there is a significant problem with category imbalance in the CD tasks. Therefore, the dice loss is introduced to reduce the effect that the sample imbalance has on the overall result. 
The dice coefficient is described as the similarity of two contour regions, and the higher the value, the more similar the contour regions are to one another. Moreover, we need to make the objective function as small as possible when we are training the network. For this reason, we subtract the dice coefficient from one to optimize as follows:
\begin{equation}
	\mathcal{L}_{dice}=1-\frac{2 \times \sum_{i=1}^{N}(\textit{t}_i\textit{p}_i)}
		{\sum_{i=1}^{N}\textit{t}_i + \sum_{i=1}^{N}\textit{p}_i}
\end{equation}

The hybrid loss function of the network is defined as follows:
\begin{equation}
	\mathcal{L} =\mathcal{L}_{bce} +\mathcal{L}_{dice}
\end{equation}

\section{Experiments and Analysis}
To investigate the performance of the proposed network, experiments have been conducted on three of the most challenging datasets, namely WHU-CD \cite{ref63}, LEVIR-CD \cite{ref54}, and SYSU-CD \cite{ref33}. In this section, each of the three datasets will be described in detail. Then, a description of the evaluation metrics and parameter settings is provided. After that, numerous SOTA CD methods are given for comparison. Finally, a comprehensive analysis is conducted of the experimental results. Meanwhile, the ablation studies are also being designed to test the performance of the AFCF and SE modules.
\subsection{Dataset Description}
\begin{enumerate}
\item{WHU-CD: The WHU-CD is a public building CD dataset. This dataset contains a pair of images with a spatial resolution of 0.2m and a size of 32507$\times$15354 pixels. Following a previous study \cite{ref64}, we first cropped the two original images into small patches of size 256$\times$256 such that there is no overlap. Then, the dataset is randomly divided into 6096, 762, and 762 for the training, validation, and test, respectively.}
\item{LEVIR-CD: The LEVIR-CD dataset includes a total of 637 image pairs with a size of 1024$\times$1024. This dataset is collected using Google Earth images with a spatial resolution of 0.5 m from 2002 to 2018. Similarly, following the division dataset in \cite{ref54}, the number of image pairs is 7120, 1024, and 2048 for the training, validation, and test, respectively.}
\item{SYSU-CD: The SYSU-CD is a newly public challenging  dataset that is published by Shi \textit{et al.} \cite{ref33}. Different from WHU-CD and LEVIR-CD, this dataset includes a wide range of different type of changes, such as road expansion, change of vegetation, and sea construction. The SYSU-CD dataset contains a total of 20000 pairs of high-resolution images with a spatial resolution of 0.5m and a size of 256$\times$256. The numbers of image pairs are 12000, 4000, and 4000 for the training, validation, and test sets, respectively.}
\end{enumerate}
\subsection{Evaluation Metrics}
The experiment results are evaluated based on four different metrics: precision (Pre), recall (R), F1-score (F1), and intersection over union (IoU). F1 and IoU are a comprehensive evaluation of the performance of the network. These metrics are defined as follows:
\begin{equation}
	Pre = \frac{TP}{TP+FP}
\end{equation}
\begin{equation}
	R = \frac{TP}{TP+TN}
\end{equation}
\begin{equation}
	Pre = 2\times \frac{Pre\times R}{Pre + R}
\end{equation}
\begin{equation}
	IoU = \frac{TP}{TP+FP+FN}
\end{equation}
where true positive (TP) represents the changed pixel numbers detected correctly, false positive (FP) denotes the pixel numbers of unchanged that are wrongly predicted as changed one, false negative (FN) denotes the pixel numbers of the changed pixel that are unpredicted, and true negative (TN) stands for the pixel numbers of unchanged pixel predicated correctly.
\subsection{Implementation Details}
The proposed method is conducted with the PyTorch framework by using an i7-10700K CPU at 3.8 GHz and 32 GB of RAM and an RTX3080Ti with 12GB of memory. The adaptive moment estimation (Adam) is used as the optimizer; its initial learning rate is stated as 1e-4, momentum is 0.9, and weight decay is stated as 1e-4. The epoch number and batch size in all methods are set to 100 and 8, respectively. In addition, following each training epoch, validation is conducted, and the best validation model is evaluated on the test sets.
\subsection{Comparison Methods}
We compare the proposed method to eight SOTA CD methods to verify the effectiveness of the proposed method. Including FC-EF \cite{ref22}, FC-Siam-Conc \cite{ref22}, FC-Siam-Diff \cite{ref22}, DSIFN\cite{ref24}, STANet \cite{ref54}, DSAMNet \cite{ref33}, SNUNet \cite{ref34}, 
ChangeFormer \cite{ChangeFormer}, BITNet \cite{ref36}, and LightCDNet \cite{LightCDNet}.
\begin{enumerate}
	\item {FC-EF: To obtain the change map, this network concatenates the bi-temporal images directly to the channel dimension as the network’s input.}
	\item {FC-Siam-Conc: Instead of directly fusing bi-temporal images into one composite image, the Siamese network is introduced into this model. To be more specific, the dual-branch features that are extracted from the Siamese network and then conducted a concatenation operation.}
	\item {FC-Siam-Diff: Its architecture is similar to the FC-Siam-Conc. The dual-branch features that are extracted from the Siamese network and then conducted a difference operation.}
	\item {DSIFN: The DSIFN adopts a Siamese network to extract dual-branch features in the encoding stage. To deal with spatial heterogeneity, the DSIFN introduced the attention modules to fuse the heterogeneous features during the decoding stage.}
	\item {The STANet is a metric-based CD method. A spatial-temporal attention module is designed to connect the relationships between any two pixels to generate more discriminating features.}
	\item {DSAMNet: The DSAMNet is also a Siamese network that integrates convolutional block attention modules into the metric module to generate more discriminating information. In addition, an auxiliary tool called the deep supervision (DS) module has been introduced to improve the feature extractor’s capacity for learning.}
	\item {SNUNet-32: The densely skip connection strategy is introduced to alleviate the information loss. Besides, this network proposes an ensemble channel attention module (CBAM) to enhance its features. The channel size of the SNUNet is set to 32 in this paper.}
	\item {ChangeFormer: A method based on Transformer to extract hierarchical features. The extracted hierarchical features are differenced and then aggregated in the decoder to obtain the CD result.}
	\item {BITNet: The BITNet is a transformer-based network that expresses bi-temporal images as a small number of model contexts and semantic tokens in a token-based space-time.}
	\item {LightCDNet: The LightCDNet uses the Siamese architecture for CD. This method designs a multi-temporal feature fusion to improve the representation of change information.}
\end{enumerate}
To establish a fair comparison, we reproduce them by utilizing their available source codes and doing the tests with the hyperparameters set at their default values.

\textit{1) Quantitative Comparison:} The overall comparison results of the extensive experimental work on the three datasets are shown in Table \ref{tableI}.
On the WHU-CD dataset, it is evident that the proposed AFCF3D-Net outperforms the other SOTA methods in all indicators. Specifically, the proposed method obtains the best F1 and IoU, which are achieved at 93.58\% and 87.93\%, respectively.
The proposed AFCF3D-Net achieves improvements of 2.08\% and 3.63\% in F1 and IoU, respectively, in comparison to the LightCDNet, which takes the ranking of second on the list. On the LEVIR-CD dataset, the proposed method achieves the highest scores on both F1 and IoU, which are presented at 90.76\% and 83.08\%, respectively. The ChangeFormer reaches the second-ranked scores of 90.40\% and 82.48\% in terms of F1 and IoU, respectively. The proposed AFCF3D-Net achieves improvements of approximately 0.36\% of F1 and 0.60\% of IoU, respectively, when compared to the second-ranked ChangeFormer. As for Precision and Recall metrics, the proposed method achieves the second-best with Precision and IoU of 91.35\% and 90.17\%, respectively. The ChangeFormer and the STANet obtained the highest Precision and Recall of 92.05\% and 91.00\%, which is 0.70\% and 0.83\% higher than that of the proposed AFCF3D-Net, respectively. On the SYSU-CD dataset, according to Table I, it is evident that the proposed AFCF3D-Net achieves optimal scores, with Recall, F1, and IoU values of 83.88\%, 83.11\%, and 71.09\%, respectively. Although the FC-Siam-conc obtains the highest Precision of 83.54\%, its Recall result of 69.91\% is not satisfactory. The above quantitative analysis leads us to the conclusion that the proposed AFCF3D-Net is superior to the other SOTA CD methods in terms of performance.

\begin{table*}
	\caption{Quantitative Performance Comparison Between The Proposed Method and The State-of-the-Art Methods on Three Datasets. All Results are Descripted using Percentages (\%).}
	\centering
		\begin{tabular}{ccccccccccccc}
			\toprule
			\multirow{2}{*}{Methods}&\multicolumn{4}{c}{\textbf{WHU-CD}}&\multicolumn{4}{c}{{\textbf{LEVIR-CD}}}&\multicolumn{4}{c}{\textbf{SYSU-CD}}\\
			\cmidrule(lr){2-5}\cmidrule(lr){6-9}\cmidrule(lr){10-13}
			&P&R&F1&IoU&P&R&F1&IoU&P&R&F1&IoU\\
			\midrule
			FC-EF&79.33&74.58&76.88&62.45&85.87&82.22&83.35&72.43&80.16
			&70.69&75.13&60.17\\
			FC-Siam-conc &48.58&85.49&61.96&44.89&86.18&85.12&87.58&77.91&
			\textcolor{Maroon}{83.54}&69.61&75.94&61.21\\
			FC-Siam-diff &67.55&63.21&65.31&48.75&88.59&80.72&85.37&74.48&
			78.34&66.13&70.17&55.11\\
			DSIFN &85.89&\textcolor{NavyBlue}{91.31}&88.52&79.40&87.3&88.57&	88.42&78.09&79.32&73.85&77.46&62.94\\
			STANet
			&86.11&	88.14&87.11&77.17&83.81&\textcolor{Maroon}{91.00}&	87.30&	77.40&73.33&\textcolor{NavyBlue}{82.73}&77.75&63.59\\
			DSAMNet
			&84.86&87.58&86.20&75.68&89.06&87.21&88.13&77.82
			&74.81&81.86&78.18&64.35\\
			SNUNet &82.63&90.33&86.31&75.92&90.55&89.28&89.91&81.67&82.16&71.33&76.36&61.76\\
			ChangeFormer &89.36&90.28&89.82&81.60&\textcolor{Maroon}{92.05}&88.80&\textcolor{NavyBlue}{90.40}&\textcolor{NavyBlue}{82.48}&77.16&78.51&77.83&63.71\\
			LightCDNet &92.00&91.00&\textcolor{NavyBlue}{91.50}&\textcolor{NavyBlue}{84.30}&91.30&88.00&89.60&81.20&\textcolor{NavyBlue}{83.01}&74.90&\textcolor{NavyBlue}{78.75}&\textcolor{NavyBlue}{64.98}\\
			BITNet
			&\textcolor{NavyBlue}{92.71}&89.83&91.25			&83.91&89.24&89.37&89.31&80.68&80.40&77.09&78.72&64.90\\
			Ours&\textcolor{Maroon}{93.47}&\textcolor{Maroon}{93.69}&\textcolor{Maroon}{93.58}
			&\textcolor{Maroon}{87.93}&\textcolor{NavyBlue}{91.35}&\textcolor{NavyBlue}{90.17}&\textcolor{Maroon}{90.76}&\textcolor{Maroon}{83.08}&82.30&\textcolor{Maroon}{83.88}&\textcolor{Maroon}{83.11}&\textcolor{Maroon}{71.09}\\
			\bottomrule
		\end{tabular}
			\begin{tablenotes}
				\centering
				\footnotesize
				\item * Red is used to highlight the \textcolor{Maroon}{best} result, and blue is used to highlight the \textcolor{NavyBlue}{second-best} result.
			\end{tablenotes}
	\label{tableI}
\end{table*}

\begin{figure*}[ht]		
	\centering
	\includegraphics[scale=0.9]{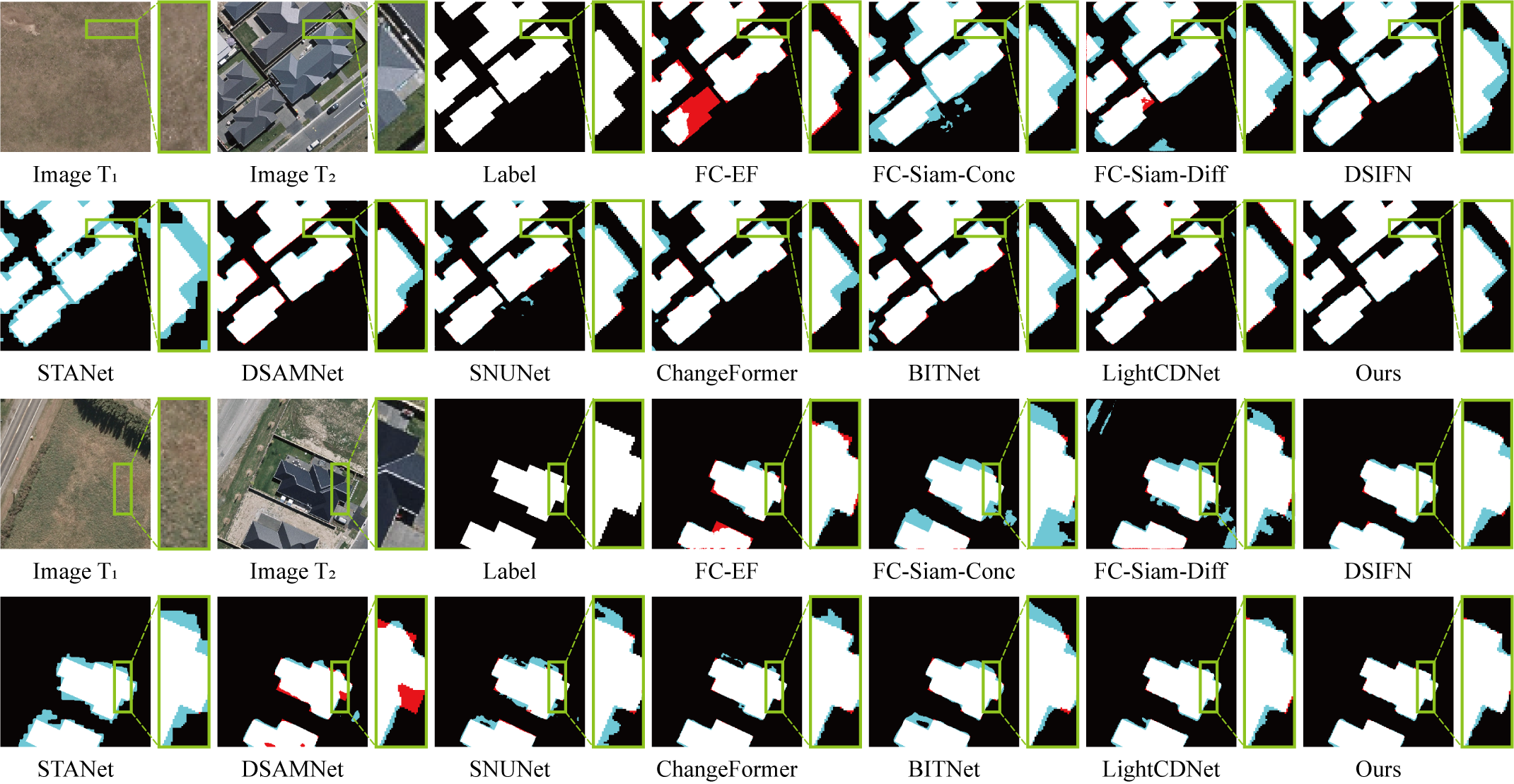}
	\caption{Visual results of different methods in test samples of the WHU-CD dataset.}
	\label{figure7}
\end{figure*}

\begin{figure*}[!h]
	\centering
	\begin{minipage}{1.0\linewidth}
		\centering
		\includegraphics[scale=0.9]{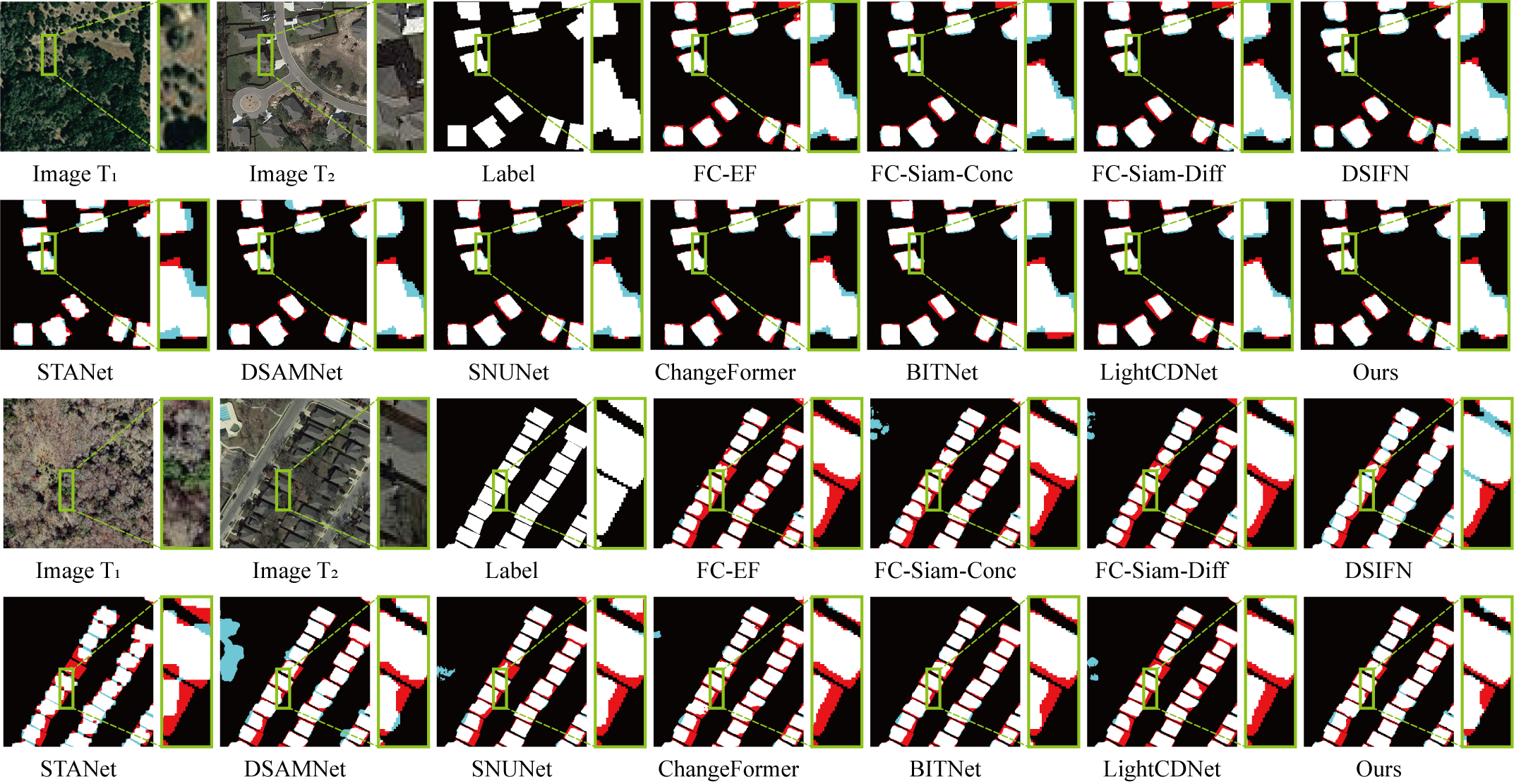}
		\caption{Visual results of different methods in test samples of the LEVIR-CD dataset.}
		\label{figure8}
		\vspace{10pt}
	\end{minipage}
	%
	\begin{minipage}{1.0\linewidth}
		\centering
		\includegraphics[scale=0.9]{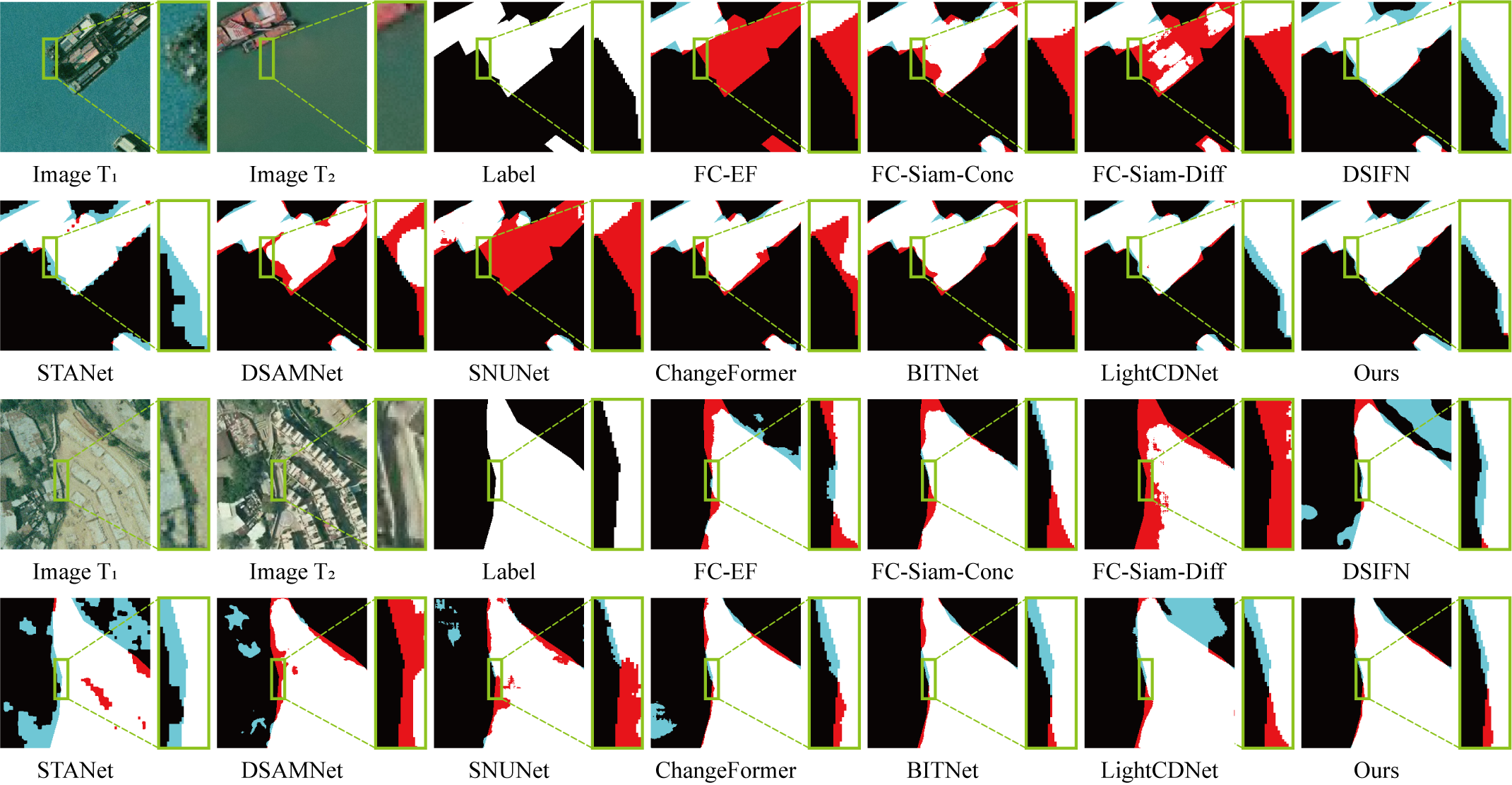}
		\caption{Visual results of different methods in test samples of the SYSU-CD dataset.}
		\label{figure9}
	\end{minipage}
\end{figure*}

\textit{2) Qualitative  Comparison:} Figs. \ref{figure7}-\ref{figure9} provide the intuitive visualization results of each method on three datasets. To better identify the differences between the change detection results and the labels, four colors are used to show the detection results: white for TP, blue for FN, red for FP, and black for TN. In addition, to have more clarity of the results of each method at the boundary, we enlarged particular areas of the change results.

The WHU-CD dataset is mainly focused on building change, and the representative changed samples from WHU-CD datasets are shown in Fig. \ref{figure7}, including the similar color appearance between the buildings and the asphalt-covered roads and the building with various shooting angles. From the first set of images in Fig. \ref{figure7} and their enlarged views, it can be clearly observed that the proposed method accurately detects building changes.
Even though the DSAMNet, BITNet and LighCDNet methods have good results, those results contain more false negatives (blue). From the second sets of images and their enlarged views, we can see that the proposed method has fewer false positives and false negatives. This means that the proposed AFCF3D-Net gets more accurate boundary details for the areas of the building that have changed.

Different from the WHU-CD dataset, more small-change buildings are contained in the LEVIR-CD dataset, and the small-change buildings make it more difficult to yield accurate results. As shown in Fig. \ref{figure8}, as for the densely and small targets of changed buildings, the FC-EF, FC-Siam-diff, and FC-Siam-conc methods obtain a higher false positive compared to the other methods. From the second set of images in Fig. 8, the results of the DSIFN and STANet are not satisfactory because the changed results stuck together. Besides, for the SNUNet, ChangeFormer, BITNet, and LightCDNet, their results have more false negatives compared with the proposed method. By contrast, as can be seen in Fig. 8, the proposed AFCF3D-Net could maintain better boundary details and achieve the best visual results among all methods.

To further test the effectiveness of the AFCF3D-Net, the SYSU-CD dataset is used to conduct an experiment. The SYSU-CD is a challenging dataset that includes diverse change situations such as road expansion, sea construction, groundwork before construction, etc. In Fig. \ref{figure9}, which displays the visualization results of two different and various changes, it is evident that the proposed AFCF3D-Net still preserves the more precise boundaries of the changed targets. False positives are a major issue in the results of the FC-EF, FC-Siam-conc, and FC-Siam-diff. As for the DSIFN and STA, their results contain much false negatives. The visual results of DSAMNet and SNU are not satisfactory due to the number of false positives as well as false negatives in both collections.
Although the ChangeFormer, BITNet, and the proposed AFCF3D-Net have better visual results compared to other methods, the proposed AFCF3D-Net has fewer falsely detected pixels in the results compared to the ChangeFormer and BITNet. In general, the proposed AFCF3D-Net achieves the best CD performance.

\textit{3) Model complexity:}
Table \ref{tableII} shows the model complexity of the proposed AFCF3D-Net from two indications: the number of their model parameters (Params.) and the floating point operations per second (FLOPs).

Among all the compared methods, the FC-EF, FC-Siam-diff, and FC-Siam-conc have relatively low levels of complexity; the reason is that their feature extraction networks are simpler than those of the other methods. The BITNet uses a simple CNN (i.e., ResNet-18) to extract features without other sophisticated structures such as FPN and UNet. As a result, the complexity of the BITNet is only marginally higher than that of the other methods, such as FC-EF and FC-Siam-diff. The SNUNet uses a simple network module to extract features, but generates the change map by using a dense skip connection strategy, which introduces more model parameters.
The LightCDNet is regarded as a lightweight network for CD, but its parameters are still 10.75 M. As for the ChangeFormer network, its parameters are the second-largest. Since the proposed method introduces the time dimension, which brings a large number of model parameters and is in the middle of the pack, it contains a series of spatial and time convolutions in terms of FLOPs.
Moreover, the sophisticated AFCF fusion module and full-scale skip connection cause the relatively large FLOPs of the proposed method. However, the AFCF3D-Net effectively improves the accuracy of results compared to the other methods.
\begin{table}[!h]
	\caption{Model Complexity Comparisons.}
	\centering
	\begin{tabular}{cccc}
		\toprule
		Methods& Params (M) & FLOPs (G)& F1(\%)\\
		\midrule
		FC-EF&1.35&3.58&76.88\\
		FC-Siam-conc&1.35&4.73&61.96\\
		FC-Siam-diff &1.55&5.33&65.31\\
		DSIFN &50.46&50.77&88.52\\
		STANet&16.89&6.43&87.11\\
		DSAMNet&16.95&75.39&86.20\\
		SNUNet&12.03&54.83&86.31\\
		ChangeFormer&
		29.75&21.18&89.82\\
		LightCDNet&10.75
		&21.54&91.50\\
		BITNet&3.04&8.75&91.25\\
		Ours&17.54&31.72&93.58\\
		\bottomrule
	\end{tabular}
	\label{tableII}
\end{table}
\subsection{Ablation Studies}
The ablation experiments are designed on three datasets to test the efficacy of the AFCF and SE modules. These experiments involve removing or combing the two modules that make up the AFCF3D-Net to discuss the efficacy of each module within the network. To begin with, the Siamese encoder and decoder without any other module is defined as “baseline”, the SE module is denoted as “SE”, the AFCF module is represented as “AFCF”, and the AFCF module without a residual connection branch is denoted as “AFCF-outR”.

\begin{figure*}[ht]		
	\centering
	\includegraphics[scale=0.9]{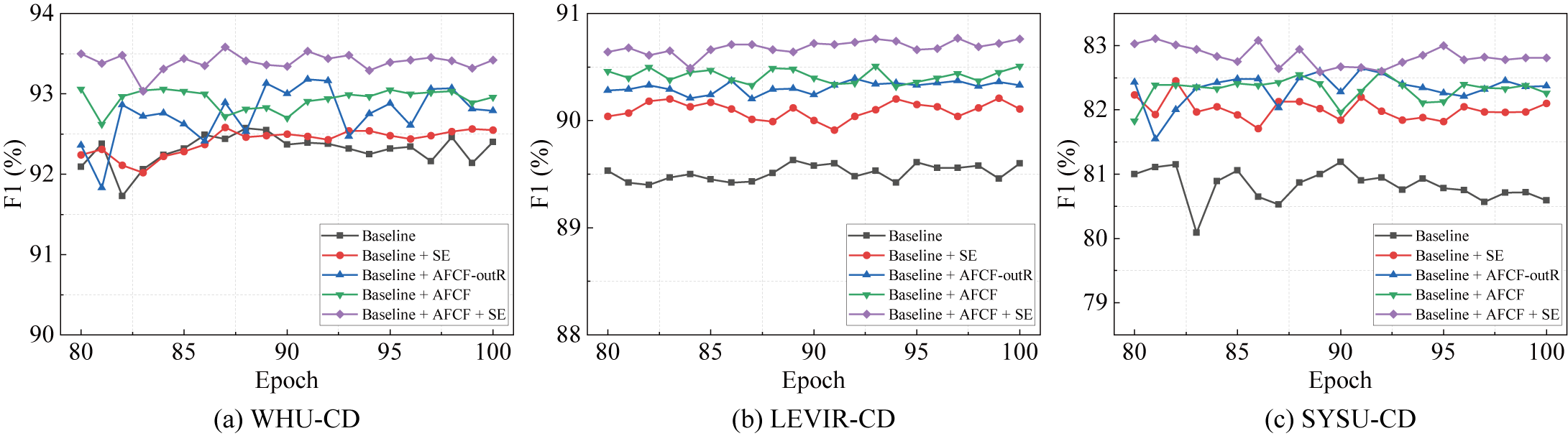}
	\caption{The results of three test sets from Epoch 80 to 100 of different network mode.}
	\label{figure10}
\end{figure*}

We mainly use F1 and IoU to compare each network mode. Fig. \ref{figure10} shows the results of the test sets from Epoch 80 to 100 of each network mode, and Table \ref{tableIII} shows the average performance of each network mode on three datasets. It is clear that the “Baseline” gets the lowest F1 and IoU. While the “SE” module is added, the results on three datasets show different degrees of improvement. Specifically, the improvements in F1 are 0.11\% and 0.58\% on the WHU-CD and LEVIR-CD, respectively. while the improvements in F1 are 1.19\% on the SYSU-CD. These results demonstrated that the “SE” module can capture more distinguishable change information. Compared to the “SE” module, the experiment results on three datasets are better when the “AFCF” module is added. Moreover, when we combine the “SE” and “AFCF-outR” modules, the results on three datasets are both better than those of the “Baseline + SE”. The reason might be that the “AFCF” module can alleviate the semantic gap in the full-scale features. However, compared to the “Baseline + AFCF”, the “Baseline + AFCF-outR” slightly decreased on three datasets. Specifically, F1 is decreased by 0.18\%, 0.11\%, and 0.02\% on the three datasets, respectively. In this case, we infer that the adjacent branch in the AFCF module is just one of the auxiliary features. After combining the “AFCF” and “SE” modules, the results have significantly advanced while also achieving the best performance of all the modules in comparison. In particular, on the WHU-CD dataset, compared to the “Baseline”, the “Baseline + AFCF + SE” can boost F1 from 92.30\% to 93.39\%. Compared to only using the “SE” or “AFCF” modules, F1 is improved by 0.98\% and 0.46\%, respectively. Fig. \ref{figure10} shows the curves representing F1 values of these methods. We can see that F1 of “Baseline” is lowest on three datasets. As for “Baseline + AFCF-outR”, its F1 has large fluctuations on the WHU-CD and SYSU-CD datasets. “Baseline + AFCF + SE” achieves the highest performance. In sum, the results show that the “AFCF” and “SE” modules can effectively improve accuracy, and that effectiveness is verified.
\begin{table*}
	\caption{Evaluation of Proposed Network with Different Mode on Three Datasets. The Best Score is Marked in Bold.}
	\centering
	\begin{tabular}{cccccccc}
		\toprule
		\multirow{2}{*}{Network mode}&\multicolumn{2}{c}{\textbf{WHU-CD}}&\multicolumn{2}{c}{{\textbf{LEVIR-CD}}}&\multicolumn{2}{c}{\textbf{SYSU-CD}}&\multirow{2}{*}{Params (M)}\\
		\cmidrule(lr){2-3}\cmidrule(lr){4-5}\cmidrule(lr){6-7}
		&F1 (\%) &IoU(\%)&F1 (\%) &IoU(\%)&F1 (\%) &IoU(\%)\\
		\midrule
		Baseline&92.30&	85.72&89.51&81.02&80.82&67.81&16.11\\
		Baseline + SE &92.41&85.89&90.09&81.97&82.01&69.49&17.34\\
		Baseline + AFCF-outR&92.75&86.80&90.31&82.33&82.32&69.95&16.27\\
		Baseline + AFCF &92.93&87.28	&90.42&82.51&82.34	&69.98&16.27\\
		Baseline + AFCF + SE
		&\textbf{93.39}&\textbf{87.56}&\textbf{90.68}	&\textbf{82.96}&\textbf{82.83}&\textbf{70.69}&	17.54\\
		\bottomrule
	\end{tabular}

	\label{tableIII}
\end{table*}

\subsection{3D-CNN Fusion VS 2D-CNN Fusion}
It is an important contribution to the proposed AFCF3D-Net that 3D convolution is used to fuse the bi-temporal features. To verify its effectiveness, we conducted the comparative experiments between the 3D-CNN fusion strategy and the 2D-CNN fusion strategy on the three datasets. In the 2D-CNN fusion network, the bi-temporal images are concatenated directly as the network input, and the architectures of the encoder and decoder are the same as in the proposed AFCF3D-Net except for using 2D convolution instead of 3D convolution. It should be noted that the input channel number of the first convolution layer in the original ResNet-18 is 3, but we changed it to 6 in our variant of the compared method, therefore, the first convolution layer does not use pre-trained parameters.

\begin{figure}[ht]		
	\centering
	\includegraphics[scale=0.8]{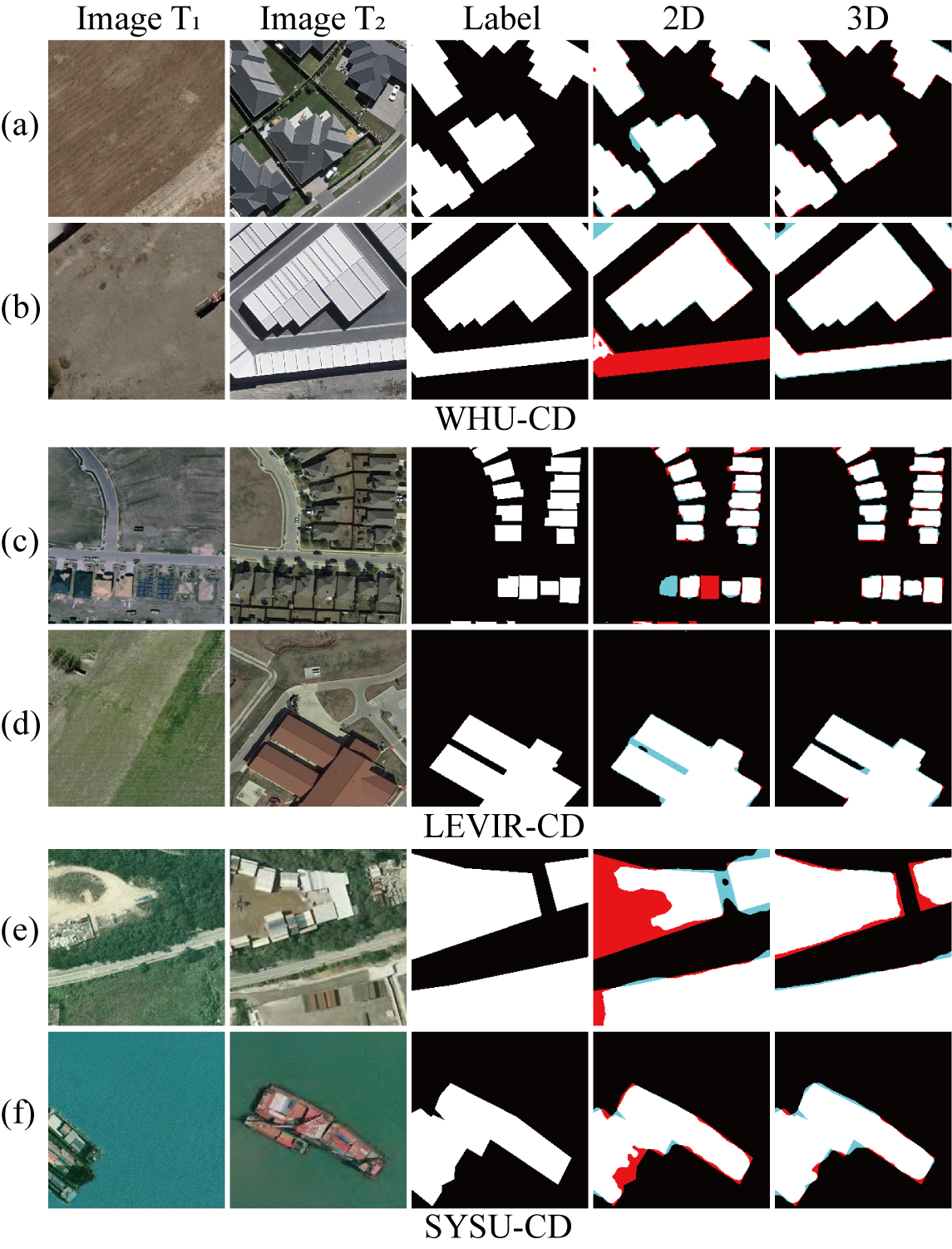}
	\caption{Visual comparison between the 2D convolution fusion and the 3D convolution fusion on three datasets.}
	\label{figure11}
\end{figure}

Table \ref{tableIV} lists the results of the two fusion strategies on three datasets. It is easy to observe that the results achieved by utilizing the 3D convolution are superior to those acquired by utilizing the 2D convolution. For example, on the WHU-CD, the 3D convolution fusion achieves improvements of approximately 2.19\% and 3.78\% of F1 and IoU, respectively, in comparison to the 2D convolution. On the LEVIR-CD, when compared to the 2D convolution fusion, the 3D convolution results in improvements of approximately 1.64\% of F1 and 2.7\% of IoU, respectively. It is worth noting that on the SYSU-CD dataset, F1 and IoU of 2D convolution are 3.49\% and 4.95\% lower than those of 3D convolution, respectively. The reason is that there are many types of changes in this dataset. Compared to the direct concatenation of 2D convolution fusion, the inner fusion property of 3D convolution can obtain more representative features. As a result, the 3D convolution obtains higher accuracy. Fig. \ref{figure11} provides a more intuitive visualization of the results on the three datasets. We can see that the proposed 3D convolution fusion achieved more accurate change information than the 2D convolution fusion.

\begin{table}
	\caption{Evaluation of Different Strategy on Three Datasets. The Best Score is Marked in Bold.}
	\centering
	\begin{tabular}{ccccccc}
		\toprule
		\multirow{2}{*}{}&\multicolumn{2}{c}{\textbf{WHU-CD}}&\multicolumn{2}{c}{{\textbf{LEVIR-CD}}}&\multicolumn{2}{c}{\textbf{SYSU-CD}}\\
		\cmidrule(lr){2-3}\cmidrule(lr){4-5}\cmidrule(lr){6-7}
		&F1 (\%) &IoU(\%)&F1 (\%) &IoU(\%)&F1 (\%) &IoU(\%)\\
		\midrule
		2D&91.39&84.15&89.12&80.38&79.62&66.14\\
		3D &\textbf{93.58}&\textbf{87.93}&\textbf{90.76}&\textbf{83.08}&\textbf{83.11}&\textbf{71.09}\\
		\bottomrule
	\end{tabular}
	\label{tableIV}
\end{table}

\section{Conclusion}
In this paper, a novel end-to-end 3D convolutional network called AFCF3D-Net is proposed for RS image CD. Instead of using the existing feature fusion strategy directly, we designed a new feature fusion strategy based on 3D convolution. To overcome the semantic gap between low-level and high-level features, we propose an AFCF module to aggregate the complementary information between the adjacent-level features. In addition, we also employ the dense skip connection strategy to enhance the capability of pixel-wise prediction for change detection. Given the approaches above, the experiments on the three challenging datasets (WHU-CD, LEVIR-CD, and SYSU-CD) demonstrate that the AFCF3D-Net outperforms the eight SOTA methods. Additionally, the main drawback of the proposed AFCF3D-Net is its high complexity. In the future, we will focus on designing a new lightweight network to improve CD results.

\section*{Acknowledgments}
The authors would like to thank everyone who has contributed datasets and fundamental research models to the public. They also appreciate the editors and anonymous reviewers for their valuable comments, which greatly improved the quality of the paper.

\balance
\bibliographystyle{IEEEtran}
\bibliography{reference}




\end{document}